\RequirePackage{fix-cm}
\documentclass[twocolumn]{svjour3}          
\smartqed  
\usepackage{graphicx,cite,soul,framed,epsfig,mathptmx,times,
amsmath,amssymb,color,flushend,gensymb,subfigure,fancyhdr,
subfigure}
\usepackage{lipsum,epstopdf}

\journalname{Journal of Intelligent \& Robotic Systems}
\begin{document}

\title{Tracking Error Learning Control for Precise Mobile Robot Path Tracking in Outdoor Environment 
\thanks{The information, data, or work presented herein was funded in part by the Advanced Research Projects Agency-Energy (ARPA-E), U.S. Department of Energy, under Award Number DE-AR0000598.}
}

\titlerunning{Tracking Error Learning Control for Precise Mobile Robot Path Tacking} 
\author{Erkan Kayacan       \and
        Girish Chowdhary
}


\institute{E. Kayacan \at
              Senseable City Laboratory and Computer Science \& Artificial Intelligence Laboratory \\
               Massachusetts Institute of Technology \\
              Tel.: +217-721-8278\\
              \email{erkank@mit.edu}           
           \and
           G. Chowdhary \at
Coordinated Science Laboratory and Distributed Autonomous Systems Laboratory \\
 University of Illinois at Urbana-Champaign
}

\date{Received: date / Accepted: date}

\maketitle

\begin{abstract}

This paper presents a Tracking-Error Learning Control (TELC) algorithm for precise mobile robot path tracking in off-road terrain. In traditional tracking error-based control approaches, feedback and feedforward controllers are designed based on the nominal model which cannot capture the uncertainties, disturbances and changing working conditions so that they cannot ensure precise path tracking performance in the outdoor environment. In TELC algorithm, the feedforward control actions are updated by using the tracking error dynamics and the plant-model mismatch problem is thus discarded. Therefore, the feedforward controller gradually eliminates the feedback controller from the control of the system once the mobile robot has been on-track. In addition to the proof of the stability, it is proven that the cost functions do not have local minima so that the coefficients in TELC algorithm guarantee that the global minimum is reached. The experimental results show that the TELC algorithm results in better path tracking performance than the traditional tracking error-based control method. The mobile robot controlled by TELC algorithm can track a target path precisely with less than $10$ cm error in off-road terrain.

\end{abstract}

\section{Introduction}

Mobile robots in off-road terrain are increasingly used in wide range of nonindustrial applications such as agriculture, search and rescue, military, forestry, mining and security \cite{erkanDeNMPC, Ming2009}. However, guidance, navigation, and control of mobile robots require advanced control methods to alleviate the effects of unmodeled surface and soil conditions (e.g., snow, sand, grass), terrain topography (e.g., side-slopes, inclines), and complex robot dynamics. In the outdoor environment, it is sometimes arduous and/or almost impossible to obtain a priori model for such effects (i) modeling of robot-terrain interactions is challenging, (ii) the soil condition is often not known ahead of time, and (iii) the identification of system parameters is a cumbersome process and must be re-carried out for different terrains \cite{6606388, erkanmodelleme,Cui2017, KayacanRSS, PanRSS}.  

The initial studies on autonomous mobile robots used traditional controllers, e.g., proportional-integral-derivative, optimal, and model predictive controllers (MPCs) \cite{Normey2001, 5152217, Huynh2017, Amer2017}. Proportional-integral-derivative controllers are convenient only for single-input-single-output systems; however, mobile robots are multi-input-multi-output systems. As alternative methods, linear quadratic regulators and linear MPCs, which can be designed readily for multi-input-multi-output systems, were proposed for autonomous navigation of mobile robots in literatur \cite{7978197, 7562522, 7468507}. Since these methods require a linear system model to be designed, tracking error-based control algorithms were developed \cite{KLANCAR2007, SKRJANC2017177} in which the system model is linearized around the target path, and the total control input is obtained by the sum of feedback and feedforward control inputs \cite{BLAZIC20111}. The feedback control law is designed based on a nominal model, which is a priori model and cannot capture all the effects of uncertainties that are summarized in the previous paragraph. Moreover, the feedforward control law is derived taking the target path and nominal model into account so that it also cannot contain the effects of uncertainties. Since tracking error-based models might not represent real-time systems behavior accurately, the traditional tracking error-based control methods cannot ensure precise path tracking performance in the outdoor environment. Therefore, it can be concluded that a prerequisite for accurate tracking performance of model-based controllers is the achievement of a precise mathematical model of the system to be controlled \cite{erkantowards}. This shows that traditional approaches are not always inherently robust. Moreover, an amplitude-saturated output feedback control approach was proposed in \cite{7725990}. In this approach, the output of the control input is limited by upper and lower bounds; however, the input rate cannot be limited. Furthermore, this approach requires to know the direct relation between the input and output, which is known in tracking error-based control methods. In this paper, since traditional tracking error-based controllers use an MPC as a feedback controller, we also use an MPC controller for a fair comparison with the previous works in literature.

To cope up with restrictions on traditional controllers, adaptive control approaches and MPC scheme with friction compensation were respectively proposed in \cite{MARTINS2008, Barreto2014}, and successful results were reported for indoor applications. However, there is no evidence that these methods work well for outdoor applications where the uncertainty is so high, and soil conditions are changing. Robust trajectory tracking error model-based predictive controller was designed for unmanned ground vehicles to overcome the limitations of the tracking error-based controllers \cite{Kayacan2016}. Although it exhibited robust control performance, it could not ensure precise path tracking performance, e.g., the tracking error was more than $20$ cm. Moreover, learning-based nonlinear MPC algorithms were proposed in which online parameters estimators were used to update the system parameters in the system model for an articulated unmanned ground vehicle \cite{erkanCeNMPC, erkanDiNMPC, Kayacan2018}. Although precise path tracking performance was reported in these studies, the required computation times is large, especially for embedded applications. Therefore, the purpose of this paper is to develop a high precision control algorithm for mobile robots, which must be computationally efficient and can learn the mobile robot dynamics by utilizing longitudinal and lateral error dynamics. Thus, the feedforward control action will be in charge of the overall control of the mobile robot in steady-state behavior. 

The main contribution of this study beyond state of the art is that a novel Tracking Error Learning Control (TELC) scheme is developed and implemented in real-time for the first time in literature.  The first contribution of this paper is that the cost functions consisting of tracking error dynamics of the mobile robot are used to update the coefficients in the feedforward control law. Hence, TELC can learn mobile robot dynamics, and the feedback control action is removed from the overall control signal when the robot is on-track. Therefore,  the model-plant mismatch problem for outdoor applications cannot deteriorate the path tracking performance in the TELC scheme. The second contribution is that the stability of the TELC algorithm is proven that the TELC algorithm is asymptotically stable. The stability analysis shows the longitudinal and lateral error dynamics converge to zero if the learning coefficients are large enough. The third contribution is that it is proven that the TELC algorithm does not have any local minima so that it can reach the global minima. Moreover, it is shown that the feedforward control actions are bounded for a finite value for the coefficients in steady-state.  Along with the theoretical results, this paper also presents path tracking-test results of the presented TELC algorithm on a mobile robot. TELC algorithm results in precise mobile robot path tracking performance when compared to the traditional tracking error-based control method. 

This paper is organized as follows: The tracking error-based model is derived in Section \ref{sec_tebsm}. The traditional tracking error-based control  method is given in Section \ref{sec_tebc}. The TELC algorithm is formulated in Section \ref{sec_tel} while the update rules for the linear and angular velocity references are respectively derived in Sections \ref{sec_tel_lv} and \ref{sec_tel_lw}, and the stability analysis is given in Section \ref{sec_stability}. Experimental results on a mobile robot are presented in Section \ref{sec_exp}. Finally, a brief conclusion of the study is given in Section \ref{sec_conc}.

\section{Tracking Error-Based System Model}\label{sec_tebsm}

\begin{figure}[t!]
  \centering
  \includegraphics[width=1\columnwidth]{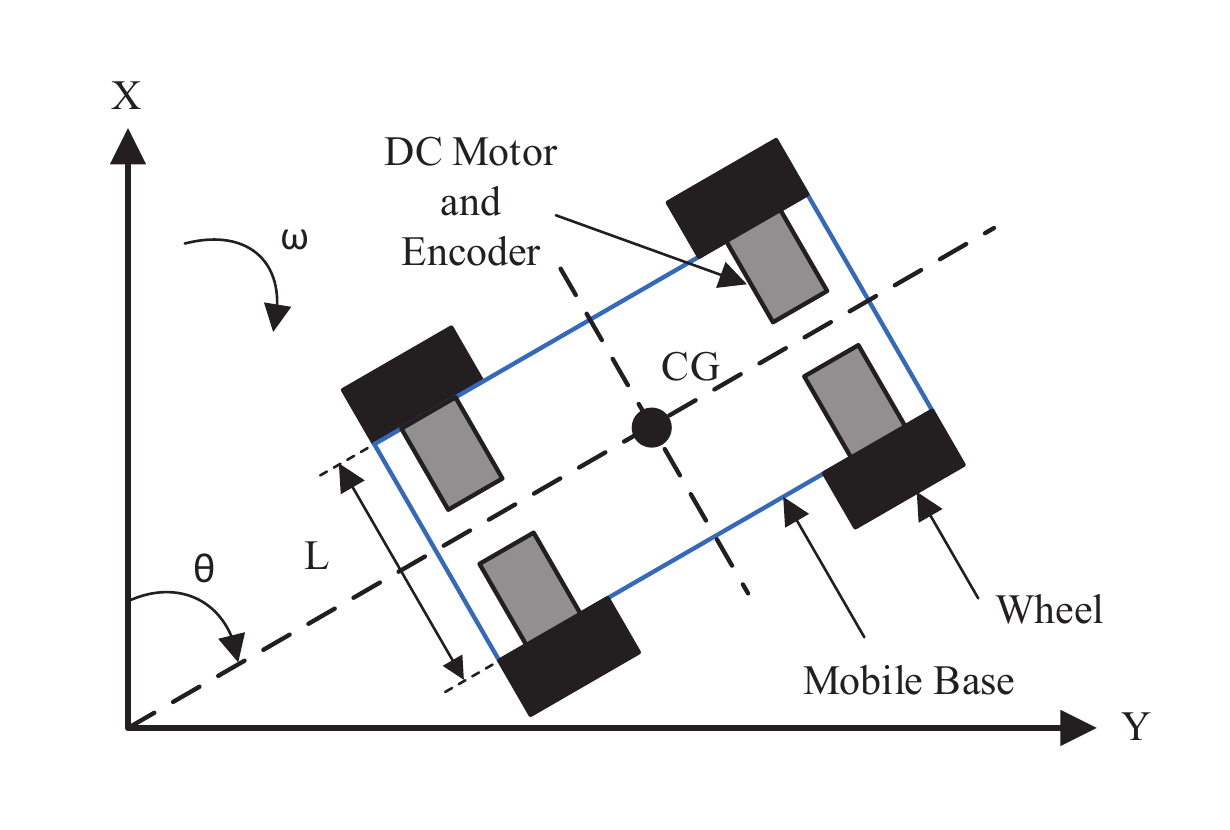}
  \caption{Schematic illustration of the mobile robot.}
  \label{fig_schematic_robot}
\end{figure}

In this paper, the mobile robot is illustrated in Fig. \ref{fig_schematic_robot}. The velocities of two driven wheels result in linear velocity $\nu=( \nu_{l}+\nu_{r} )/2$ and angular velocity $\omega= (\nu_{l} - \nu_{r})/L$ with the distance between wheels $L$, which are two control inputs of the mobile robot, $\textbf{u}=[\nu, \omega]$. The traditional unicycle model is written for a mobile robot as follows:
\begin{eqnarray}\label{eq_kinematicmodel}
 \left[
  \begin{array}{c}
   \dot{x} \\ \dot{y} \\ \dot{\theta} 
  \end{array}
  \right] =
   \left[
  \begin{array}{c}
   \nu \cos{(\theta)} \\   \nu \sin{(\theta)} \\ \omega 
  \end{array}
  \right]
\end{eqnarray}
where $x$ and $y$ are the position, $\theta$ is the heading angle, $\nu$ is the linear velocity, $\omega$ is the angular velocity of the mobile robot.

The target path with respect to the inertial reference frame fixed to the motion ground is defined by a reference state vector $\textbf{q}_{r} = (x_{r}, y_{r}, \theta_{r})^{T}$ and a reference control vector $\textbf{u}_{r} = (\nu_{r}, \omega_{r})^{T}$. The error state $\textbf{e}=[e_{1}, e_{2}, e_{3}]^{T}$ expressed in the frames on the mobile robot is written as:
\begin{equation}\label{eq_errorstate}
\textbf{e}=  \textbf{T} (\theta) \times [ \textbf{q}_{r} - \textbf{q} ]
\end{equation}
where $\textbf{T} (\theta)$ is the transformation matrix formulated as below:
\begin{equation}
\textbf{T} (\theta)= 
 \left[
\begin{array}{ccc}
\cos{(\theta)} & \sin{(\theta)} & 0 \\
-\sin{(\theta)} & \cos{(\theta)} & 0  \\
0 & 0 & 1 \\
\end{array}
 \right] \nonumber
\end{equation}

The tracking error-based model is derived by taking the time-derivative of the error state in \eqref{eq_errorstate} and the unicycle model in \eqref{eq_kinematicmodel} into account as follows:
\begin{eqnarray}\label{eq_errormodel1}
\dot{e}_{1}  & = & \omega e_{2} - \nu + \nu_{r} \cos{(e_{3})} \nonumber \\
\dot{e}_{2} & = & - \omega e_{1} + \nu_{r} \sin{(e_{3})} \nonumber \\
\dot{e}_{3} & = & \omega_{r} - \omega  
\end{eqnarray}
where $e_{1}$ is the longitudinal error, $e_{2}$ is the lateral error and $e_{3}$ is the heading angle error.

The tracking error-based model \eqref{eq_errormodel1} is linearized around the target path ($e_{1}=e_{2}=e_{3}=0$ ) as follows:
\begin{eqnarray}\label{eq_errormodel2}
\dot{e}_{1}  & = & \omega_{r} e_{2} - \nu + \nu_{r}  \nonumber \\
\dot{e}_{2} & = & - \omega_{r} e_{1} + \nu_{r} e_{3}\nonumber \\
\dot{e}_{3} & = & \omega_{r} - \omega  
\end{eqnarray}
Finally, it can be written in the state-space form as follows:
\begin{eqnarray}\label{eq_errormodel_ss}
\dot{\textbf{e}} & = & \textbf{A} \textbf{e} + \textbf{B} \textbf{u}_{e} \nonumber \\
\dot{\textbf{e}} & = & \left[
                \begin{array}{ccc}
                  0 & \omega_{r} & 0   \\
                  -\omega_{r} & 0 & v_{r} \\
                  0 & 0 & 0 \\
                \end{array}
              \right] \textbf{e}  +
              \left[ \begin{array}{cc}
                  -1 & 0   \\
                  0 & 0  \\
                 0 & -1  \\
                \end{array}\right] \textbf{u}_{e}
\end{eqnarray}
where the state and control vectors are written as
\begin{eqnarray}\label{eq_state}
\textbf{e} & = & \left[
  \begin{array}{ccc}
   e_{1} & e_{2} & e_{3}
  \end{array}
  \right]^{T} \\ \label{eq_input}
\textbf{u}_{e} & = & \left[
  \begin{array}{cc}
   \nu_{e} & \omega_{e} 
  \end{array}
  \right]^T
\end{eqnarray}
where $\nu_{e} = \nu - \nu_{r}$ and $\omega_{e}= \omega - \omega_{r}$.

\begin{remark}
The tracking error-based system model is fully controllable when either the linear velocity reference $\nu_{r}$ or the angular velocity reference $\omega_{r}$ is nonzero, which is a sufficient condition.
\end{remark}

\section{Traditional Tracking Error-Based  Control} \label{sec_tebc}

The traditional tracking-error-based control algorithm for a mobile robot is formulated in this section. The total control input applied to the mobile robot is calculated as the summation of the feedback control action $\textbf{u}_{b}$ and the feedforward control action $\textbf{u}_{f}$ as follows:
\begin{equation}\label{eq_totalinput}
\textbf{u }=  \textbf{u}_{b} + \textbf{u}_{f}
\end{equation}
The feedback controller generates the differences between the actual and reference control inputs while the feedforward control actions are the reference control inputs. The traditional feedback and feedforward control actions are formulated in following subsections.

\subsection{Feedback Control Action: Model Predictive Control}\label{sec_mpc}

A mobile robot can be described by a linear continuous-time model:
\begin{equation}
\dot{\textbf{e}} (t)= \textbf{A} \textbf{e} (t) + \textbf{B} \textbf{u}_{e} (t)
\end{equation}
where $\textbf{e}$ $\in$ $\mathbb{R}^{3}$ is the state vector and $\textbf{u}_{e} (k)$ $\in$ $\mathbb{R}^{2}$ is the control input vector. The matrices $\textbf{A}$ and $\textbf{B}$ are found from the tracking error-based system model in \eqref{eq_errormodel_ss}.

The input constraints for the mobile robot are defined for all $t \geq 0$ as follows:
\begin{eqnarray}\label{constraints}
-0.1 \; \text{m/s}  \leq &  \nu_{e} (t) & \leq  0.1 \; \text{m/s}   \\
-0.1 \; \text{rad/s}  \leq &  \omega_{e}  (t) & \leq  0.1 \; \text{rad/s} 
\end{eqnarray}

The cost function is written as follows:
\begin{equation}\label{costfunction}
J\big(\Delta \textbf{U}, \textbf{e} \big)=\frac{1}{2} \Big\{ \sum_{i=k+1}^{k+N_{p}}  \| \textbf{e} (t_{i}) \|^{2}_{Q} +  \sum_{i=k+1}^{k+N_{c}}  \| \Delta \textbf{u}_{e} (t_{i}) \|^{2}_{R}  \Big\} 
\end{equation}
where $N_p=20$ and $N_c=5$ represent the prediction and control horizons, $\triangle \textbf{u}_{e}$ is the input change, and \\
$\Delta \textbf{U} = [\Delta \textbf{u}^{T} _{e}(t_{k}),...,\Delta \textbf{u}^{T} _{e}({t_{k+N_{c}}})]^{T}$ is the matrix of the input vectors from sampling instant $t_{k}$ to sampling instant $t_{k+N_{c}}$. Since the sampling time of the experiments is equal to $200$ millisecond, the prediction and control horizons are respectively equal to $4$ second and $1$ second. The positive-definite weighting matrices $Q^{3 \times 3}$ and $R^{2 \times 2}$ are defined as follows:
\begin{equation}\label{}
Q  =  diag(1,1,1) \quad \text{and} \quad R  =  diag(1,1)
\end{equation}

The following plant objective function is solved at each sampling time for the MPC:
\begin{equation}
 \begin{aligned}
 & \underset{\textbf{e}(.), \textbf{u}_{e}(.)}{\text{min}}
 & & \frac{1}{2} \Big\{ \sum_{i=k+1}^{k+N_{p}}  \| \textbf{e} (t_{i}) \|^{2}_{Q} +  \sum_{i=k+1}^{k+N_{c}}  \| \Delta \textbf{u}_{e} (t_{i}) \|^{2}_{R}  \Big\} \\
 & \text{subject to}
  && \textbf{e}(t_{k}) =\hat{\textbf{e}}(t_{k}) \\
 &&& \dot{\textbf{e}}(t) = \textbf{A} \textbf{e}(t) + \textbf{B} \textbf{u}_{e}(t) \quad  t \in [t_{k+1}, t_{k+N_{p}}] \\
 &&& -0.1   \leq   \nu_{e}(t)  \leq  0.1  \quad  t \in [t_{k+1}, t_{k+N_{c}}] \\
 &&& -0.1   \leq   \omega_{e}(t)  \leq  0.1   \quad  t \in [t_{k+1}, t_{k+N_{c}}] 
  \end{aligned}
  \label{mpc}
\end{equation}

The convex optimization problem in \eqref{mpc} is solved for the current error state information $\hat{\textbf{e}}(t_{k})$ in a receding horizon fashion. The steps for the implementation of the MPC algorithm are summarized as below:
\begin{enumerate}
  \item Measure or estimate the current error states $\hat{\textbf{e}}(t)$.
  \item Obtain $\Delta \textbf{U}^*=[\Delta \textbf{u}^*_{e} (t_{k+1}),\ldots,\Delta \textbf{u}^*_{e} (t_{k+N_{c}})]^{T}$ by solving the optimization problem in \eqref{mpc}.
  \item Calculate the feedback control action $\textbf{u}^*_{e} (t_{k+1}) = \Delta \textbf{u}^*_{e} (t_{k+1}) + \textbf{u}^*_{e} (t_{k}) $
\end{enumerate}
The MPC problem is thereafter solved for the next sampling instant over shifted prediction and control horizons. The control input generated by the MPC $\textbf{u}^{*}_{e}$ is the feedback control action $\textbf{u}_{b}$:
\begin{equation}\label{ue}
\textbf{u}_{b}= \textbf{u}^*_{e}
\end{equation}

In tracking error control scheme, the designed MPC minimizes the tracking error between the target path and the measured position of the mobile robot, and finds the differences between the actual and reference control inputs. Therefore, the feedback control inputs generated by the MPC are not the actual control inputs, which are sent to the mobile robot. We will formulate traditional feedforward control actions in the next subsection \ref{sec_feedforward}.

\subsection{Feedforward Control Action}\label{sec_feedforward}

In open-loop control, the feedforward control laws can only drive a mobile robot on a target path if there exist no uncertainties and initial state errors. Feedforward control inputs $\nu_{r}$ and $\omega_{r}$ are derived for a given target path ($x_{r}$, $y_{r}$)  by using the unicycle model \eqref{eq_kinematicmodel}. 

The linear velocity reference, i.e., $\nu_{r}$, for a mobile robot is derived for a target path $(x_{r}, y_{r})$ defined in a time interval $t \in [0,T]$ as follows:
\begin{equation}\label{eq_ref_lv}
\nu_{r} = \pm \; \sqrt[]{ (\dot{x}_{r})^{2} + (\dot{y}_{r})^{2} } 
\end{equation}
where the sign $\pm$ is the desired driving direction of the mobile robot ($+$ for forward, $-$ for reverse). 

The heading angle reference is derived from \eqref{eq_kinematicmodel} as follows:
\begin{equation}\label{eq_ref_ha}
\theta_{r} = \arctan2{(y_{r}, x_{r})} + \gamma \pi
\end{equation}
where $\gamma=0, 1$ is the desired driving direction of the mobile robot ($0$ for forward, $1$ for reverse) and the function $\arctan2$ is a four-quadrant inverse tangent function. The angular velocity reference, $\omega_{r}$, for a mobile robot is derived by taking the time-derivative of \eqref{eq_ref_ha} for a given target path $(x_{r}, y_{r})$ defined in a time interval $t \in [0,T]$ as follows:
\begin{equation}\label{eq_ref_av}
\omega_{r} = \frac {\dot{x}_{r} \ddot{y}_{r} - \dot{y}_{r} \ddot{x}_{r} } {(\dot{x}_{r})^{2} + (\dot{y}_{r})^{2}} 
\end{equation}

\begin{remark}
The necessary condition in the path generation is that the target path must be twice-differentiable, and the linear velocity reference must be nonzero, i.e., $\nu_{r}\neq0$.
\end{remark}

\section{Tracking Error Learning Control  }\label{sec_tel}

The tracking error-based model is linearized around a target path by assuming that the longitudinal, lateral and heading angle errors are around zero. Therefore, the mismatch between the tracking error-based model and real system might result in unsatisfactory control performance when the mobile robot is not on-track. Moreover, there always exist uncertainties and unmodeled dynamics, which cannot be modeled, in outdoor applications.

In the TELC, the longitudinal and lateral error dynamics are used to train the learning algorithm, which is required to learn the uncertainties and unmodelled dynamics and to keep the system on track. The learning algorithm generates the references for the control inputs. In other words, it learns the dynamics of the real system.

\begin{figure*}[t!]
  \centering
\includegraphics[width=1.6\columnwidth]{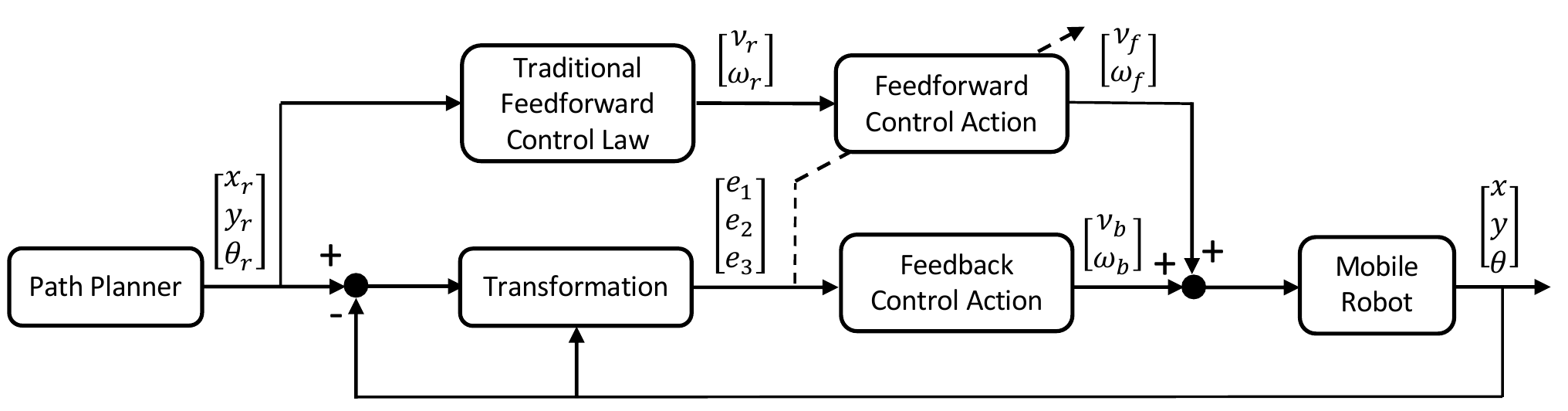}
  \caption{Tracking error learning control structure for a mobile robot.}\label{fig_tel_diagram}
\end{figure*}

Like \eqref{eq_totalinput}, the control inputs consisting of the feedback and feedforward control actions are written as follows:
\begin{eqnarray}\label{eq_v}
\nu &=& \nu_{b} + \nu_{f} \\ \label{eq_omega}
\omega &=& \omega_{b} + \omega_{f}
\end{eqnarray}
where $\nu_{b}$ and $\omega_{b}$ are the traditional feedback control actions formulated in Section \ref{sec_mpc}, while $\nu_{f}$ and $\omega_{f}$ are new feedforward control actions that are updated by the tracking error learning algorithm.  The new feedforward control actions are formulated as follows:
\begin{eqnarray}\label{eq_vf}
\nu_{f} &=& \nu_{r} k_{\nu,1}  + k_{\nu,0}  \\ \label{eq_omegaf}
\omega_{f} &=& \omega_{r} k_{\omega,1}  + k_{\omega,0} 
\end{eqnarray}
where $\nu_{r}$ and  $\omega_{r}$ are the traditional feedforward control actions formulated in \eqref{eq_ref_lv} and \eqref{eq_ref_av}, respectively. $k_{\nu,1}$ and $k_{\nu,0}$ are the coefficients to update the linear velocity reference and $k_{\omega,1}$ and $k_{\omega,0}$ are the coefficients to update the angular velocity reference. TELC structure for a mobile robot is shown in Fig. \ref{fig_tel_diagram}. In the next subsection, we will formulate the update rules for these coefficients. 

\subsection{Linear Velocity}\label{sec_tel_lv}

The requirement for the Lyapunov stability, i.e., $e_{1}=0$, is satisfied for the selection of the coefficients for the linear velocity in \eqref{eq_vf}. As can be seen from \eqref{eq_errormodel_ss}, the linear velocity appears in the channel of the longitudinal error $e_{1}$. Therefore, we use the following squared first-order longitudinal error dynamics as the cost function:
\begin{equation}\label{eq_lv_ef}
E_{\nu}= \frac{1}{2} (\dot{e}_{1} + \lambda_{\nu} e_{1} )^{2} 
\end{equation}
where $\lambda_{\nu}$ is a positive constant, i.e., $\lambda_{\nu} > 0$. It is implied that if the cost function converges to zero, i.e., $E_{\nu} = 0$, then the robust control performance condition $\dot{e}_{1} + \lambda_{v} e_{1}=0$ is satisfied so that the longitudinal error converges to zero.

Gradient descent, which is a first-order iterative optimization algorithm for finding the minimum of a function, is used to minimize the cost function $E_{\nu}$. In this approach, steps are taken proportional to the negative of the gradient of the closed-loop error function, i.e., $ E_{\nu}$, to find the minimum of the cost function.  The cost function is minimized to decide the coefficients in \eqref{eq_vf} as follows:
\begin{equation}\label{eq_lv_k1}
\dot{k}_{\nu,1}  = - \alpha_{\nu} \frac{\partial E_{\nu}}{\partial  k_{\nu,1} } 
\end{equation}
where $\alpha_{\nu}$ is the learning coefficient and positive, i.e., $\alpha_{\nu} > 0$. It is re-written by using the chain rule
\begin{equation}\label{eq_lv_k1_2}
\dot{k}_{\nu,1}  = - \alpha_{\nu} \frac{\partial E_{\nu}}{\partial \nu} \frac{\partial \nu}{\partial k_{\nu,1} } 
\end{equation}
Equation \eqref{eq_lv_ef} is inserted into the equation above, it is then obtained as
\begin{equation}\label{eq_lv_k1_3}
\dot{k}_{\nu,1}   = - \alpha_{\nu}  (\dot{e}_{1} + \lambda_{\nu} e_{1} ) \frac{\partial  (\dot{e}_{1} + \lambda_{\nu} e_{1} )}{\partial \nu}  \frac{\partial \nu}{\partial k_{\nu,1} } 
\end{equation}
Considering \eqref{eq_errormodel2}, $\frac{\partial  (\dot{e}_{1} + \lambda_{\nu} e_{1} )}{\partial \nu} =-1$ is obtained and inserted into the equation above, it is obtained as
\begin{equation}\label{eq_lv_k1_4}
\dot{k}_{\nu,1}   = \alpha_{\nu}  (\dot{e}_{1} + \lambda_{\nu} e_{1} )\frac{\partial \nu}{\partial k_{\nu,1} } 
\end{equation}
If total control input for the linear velocity applied to the mobile robot \eqref{eq_v} and the feedforward control action for the linear velocity  \eqref{eq_vf} are inserted into \eqref{eq_lv_k1_4}, the adaptation of the coefficient for the linear velocity is written as follows:
\begin{eqnarray}\label{eq_lv_k1_5}
\dot{k}_{\nu,1} & = & \alpha_{\nu}  (\dot{e}_{1} + \lambda_{\nu} e_{1} ) \underbrace{\frac{\partial (\nu_{b} + \nu_{r} k_{\nu,1} + k_{\nu,0} )}{\partial k_{\nu,1}} }_{\nu_{r}}  \nonumber \\
\dot{k}_{\nu,1} & = & \alpha_{\nu}  \nu_{r} (\dot{e}_{1} + \lambda_{\nu} e_{1} )
\end{eqnarray}

The bias term $k_{\nu,0}$ for the linear velocity is computed by using the same procedure and found as:
\begin{equation}\label{eq_lv_k0}
\dot{k}_{\nu,0} =  \alpha_{\nu} (\dot{e}_{1} + \lambda_{\nu} e_{1} )
\end{equation}

\subsection{Angular Velocity}\label{sec_tel_lw}

First we take the time-derivative of the lateral error $e_{2}$ so that the angular velocity $\omega$ can appear in the same channel with the lateral error $e_{2}$. It is obtained considering \eqref{eq_errormodel2} as follows:
\begin{equation}\label{eq_lateralerror2ndorder}
\ddot{e}_{2} =- ( \omega_{r} )^{2} e_{2} +  \omega_{r} \nu -  \nu_{r} \omega 
\end{equation}

The requirement for the Lyapunov stability, i.e., $e_{2}=0$, is satisfied for the selection of the coefficients for the angular velocity in \eqref{eq_omegaf}. Therefore, we use the following squared second-order lateral error dynamics as the same cost function as follows:
\begin{equation}\label{eq_lw_ef}
E_{\omega}= \frac{1}{2} (\ddot{e}_{2} + 2 \lambda_{\omega} \dot{e}_{2} + \lambda^{2}_{\omega} e_{2}  )^{2} 
\end{equation}
where $\lambda_{\omega}$ is a positive constant, i.e., $\lambda_{\omega} > 0$. It is implied that if the cost function converges to zero, i.e., $E_{\omega} = 0$, then the robust control performance condition $\ddot{e}_{2} + 2 \lambda_{\omega} \dot{e}_{2} + \lambda^{2}_{\omega} e_{2}  =0$ is satisfied so that the lateral error converges to zero.

As explained in Section \ref{sec_tel_lv}, the gradient descent method is used to find the minimum of the cost function $E_{\omega}$ and to decide the coefficients in \eqref{eq_omegaf} as follows:

\begin{equation}\label{eq_lw_k1}
\dot{k}_{\omega,1}  = - \alpha_{\omega} \frac{\partial E_{\omega}}{\partial  k_{\omega,1} } 
\end{equation}
where $\alpha_{\omega}$ is the learning coefficient and positive, i.e., $\alpha_{\omega} > 0$. It is re-written by using the chain rule
\begin{equation}\label{eq_lw_k1_2}
\dot{k}_{\omega,1}  = - \alpha_{\omega} \frac{\partial E_{\omega}}{\partial \omega} \frac{\partial \omega}{\partial k_{\omega,1} } 
\end{equation}
Equation \eqref{eq_lw_ef} is inserted into the equation above, it is then obtained as
\begin{equation}\label{eq_lw_k1_3}
\dot{k}_{\omega,1}   = - \alpha_{\omega}  (\ddot{e}_{2} + 2 \lambda_{\omega} \dot{e}_{2} + \lambda^{2}_{\omega} e_{2} ) \frac{\partial  (\ddot{e}_{2} + 2 \lambda_{\omega} \dot{e}_{2} + \lambda^{2}_{\omega} e_{2} )}{\partial \omega}  \frac{\partial \omega}{\partial k_{\omega,1} } 
\end{equation}
Considering \eqref{eq_errormodel2} and \eqref{eq_lateralerror2ndorder}, $\frac{\partial  (\ddot{e}_{2} + 2 \lambda_{\omega} \dot{e}_{2} + \lambda^{2}_{\omega} e_{2})}{\partial \omega} =-\nu_{r}$ is obtained and inserted into the equation above, it is obtained as
\begin{equation}\label{eq_lw_k1_4}
\dot{k}_{\omega,1}   = \alpha_{\omega} \nu_{r}  ( \ddot{e}_{2} + 2 \lambda_{\omega} \dot{e}_{2} + \lambda_{\omega}^{2} e_{2})\frac{\partial \omega}{\partial k_{\omega,1} } 
\end{equation}
If total control input for the angular velocity applied to the mobile robot \eqref{eq_omega} is inserted into \eqref{eq_lw_k1_4} considering the feedforward control action for the angular velocity \eqref{eq_omegaf}, the adaptation of the coefficient for the angular velocity is written as follows:
\begin{eqnarray}\label{eq_lw_k1_5}
\dot{k}_{\omega,1} & = & \alpha_{\omega} \nu_{r} ( \ddot{e}_{2} + 2 \lambda_{\omega} \dot{e}_{2} + \lambda_{\omega}^{2} e_{2} ) \underbrace{ \frac{\partial ( \omega_{b} + \omega_{r} k_{\omega,1} + k_{\omega,0} )}{\partial k_{\omega,1} } }_{\omega_{r}} \nonumber \\
\dot{k}_{\omega,1} & = & \alpha_{\omega} \nu_{r} \omega_{r} ( \ddot{e}_{2} + 2 \lambda_{\omega} \dot{e}_{2} + \lambda_{\omega}^{2} e_{2} )
\end{eqnarray}

The bias term $\dot{k}_{\omega,0}$ for the angular velocity is computed by using the same procedure and found as:
\begin{equation}\label{eq_lw_k0}
\dot{k}_{\omega,0} =  \alpha_{\omega} \nu_{r} ( \ddot{e}_{2} + 2 \lambda \dot{e}_{2} + \lambda^{2} e_{2})
\end{equation}

\subsection{Stability Analysis}\label{sec_stability}

The summation of the cost functions used for the linear and angular velocities is used to formulate the Lyapunov function as follows:
\begin{eqnarray}\label{eq_lyapunov}
V & = &  E_{\nu} +  E_{\omega}
\end{eqnarray}
The Lyapunov function is positive semi-definite, i.e., $V \geq 0$. To check the stability of the tracking-error learning algorithm, the time-derivative of the Lyapunov function above is taken as follows:
\begin{equation}\label{eq_lyapunov_derivative}
\dot{V} = \frac{\partial E_{\nu}}{\partial t} + \frac{\partial E_{\omega}}{\partial  t}
\end{equation}
It is re-written by using the chain rule as follows:
\begin{eqnarray}\label{eq_lyapunov_derivative}
\dot{V} &=&  \frac{\partial E_{\nu}}{\partial  k_{\nu,1} } \frac{\partial  k_{\nu,1}}{\partial  t } +  \frac{\partial E_{\nu}}{\partial  k_{\nu,0} } \frac{\partial  k_{\nu,0}}{\partial  t }  \nonumber \\ 
&&+  \frac{\partial E_{\omega}}{\partial  k_{\omega,1} } \frac{\partial  k_{\omega,1}}{\partial  t } +  \frac{\partial E_{\omega}}{\partial  k_{\omega,0} } \frac{\partial  k_{\omega,0}}{\partial  t } + g(\gamma)
\end{eqnarray}
where, $g(\gamma)$ represents the derivative of the Lyapunov function $V$ with respect to the variables other than the coefficients in the formulations of the linear and angular velocities.

The time-derivatives of the coefficienting terms in \eqref{eq_lv_k1}, \eqref{eq_lv_k0}, \eqref{eq_lw_k1} and \eqref{eq_lw_k0} are inserted into the aforementioned equation, then the time-derivative of the Lyapunov function is obtained as follows:
\begin{eqnarray}\label{eq_lyapunov_derivative2}
\dot{V} &=& - \alpha_{\nu} \Big[ 
\underbrace{ ( \frac{\partial E_{\nu}}{\partial  k_{\nu,1} } )^{2} }_{\geq 0 } +  \underbrace{ (\frac{\partial E_{\nu}}{\partial  k_{\nu,0} } )^{2} }_{\geq 0} \Big]  \nonumber \\
&& - \alpha_{w} \Big[ 
\underbrace{ (\frac{\partial E_{\omega}}{\partial  k_{\omega,1} } )^{2} }_{\geq 0} +  
\underbrace{ (\frac{\partial E_{\omega}}{\partial  k_{\omega,0} } )^{2} }_{\geq 0} \Big] + g(\gamma)
\end{eqnarray}
If the learning coefficients for the linear and angular velocity references are large enough, the time-derivative of the Lyapunov function is negative, i.e., $\dot{V}<0$. This implies asymptotically stability of the learning algorithm. 

\subsection{Global Minima}

The most important concern in the tracking error learning algorithm is that the system might reach some local minima and stay in these local minima. In this section, we will show that there are no local minima for the formulation of the tracking-error learning algorithm. If the second derivatives of the cost functions with respect to variables have the same sign, then the cost functions do not have a change in the curvature sign through the variables. This implies that the cost functions do not have local minima through these variables. 

\subsubsection{Linear Velocity}
If we take the second derivative of the cost function $E_{\nu}$ for the linear velocity with respect to $k_{\nu,1}$, it is obtained as follows:
\begin{equation}\label{eq_lv_k1_gm}
 \frac{\partial^{2} E_{\nu}}{\partial  k_{\nu,1}^{2} } = - \alpha_{\nu}  v_{r}  \frac{\partial (\dot{e}_{1} + \lambda_{\nu} e_{1} )}{\partial  k_{\nu,1} } 
\end{equation}
The chain rule is applied to the equation above
\begin{equation}\label{eq_lv_k1_gm2}
 \frac{\partial^{2} E_{\nu}}{\partial  k_{\nu,1}^{2} } = - \alpha_{\nu}  \nu_{r} \frac{\partial (\dot{e}_{1} + \lambda_{\nu} e_{1} )} {\partial \nu} \frac{\partial \nu}{\partial  k_{\nu,1} } 
\end{equation}

First $\frac{\partial  (\dot{e}_{1} + \lambda_{\nu} e_{1} )}{\partial \nu} =-1$ is obtained considering \eqref{eq_errormodel2} and inserted into the equation above. Also, total control input for the linear velocity applied to the mobile robot \eqref{eq_v} and the feedforward control action for the linear velocity  \eqref{eq_vf} are inserted into the equation above. Then, \eqref{eq_lv_k1_gm2} is obtained as follows:
\begin{eqnarray}\label{eq_lv_k1_gm3}
 \frac{\partial^{2} E_{\nu}}{\partial  k_{\nu,1}^{2} } &=& \alpha_{\nu}  \nu_{r}  \underbrace{\frac{\partial (\nu_{b} + \nu_{r} k_{\nu,1} + k_{\nu,0} )}{\partial  k_{\nu,1} } }_{\nu_{r}} \nonumber \\
 \frac{\partial^{2} E_{\nu}}{\partial  k_{\nu,1}^{2} } &=& \alpha_{\nu}  (\nu_{r})^{2}  
\end{eqnarray}

The second derivative of the cost function with respect to the bias coefficient $k_{\nu,0}$ is computed by using the same procedure as follows:
\begin{eqnarray}\label{eq_lv_ko_gm}
\frac{\partial^{2} E_{\nu}}{\partial  k_{\nu,0}^{2} }  = \alpha_{\nu}
\end{eqnarray}

Equations \eqref{eq_lv_k1_gm3} and \eqref{eq_lv_ko_gm} show that the sign of the curvature of the cost function for the linear velocity \eqref{eq_lv_ef}  is always positive; therefore, there are no local minima, which indicate that the system reaches to the global minimum. After reaching the global minimum, since $\alpha_{v}$ is a constant and $v_{r}$ is bounded, the coefficient update algorithms \eqref{eq_lv_k1_5} and \eqref{eq_lv_k0} show that coefficients converge to a finite value. A finite value for the coefficients in steady-state results in a bounded feedforward control action \eqref{eq_vf}. 

\subsubsection{Angular Velocity}

If we take the second derivative of the cost function $E_{\omega}$ for the angular velocity with respect to $k_{\omega,1}$, it is obtained as follows:
\begin{equation}\label{eq_lw_k1_gm}
 \frac{\partial^{2} E_{\omega}}{\partial  k_{\omega,1}^{2} } = - \alpha_{\omega} \nu_{r} \omega_{r} \frac{\partial ( \ddot{e}_{2} + 2 \lambda_{\omega} \dot{e}_{2} + \lambda_{\omega}^{2} e_{2} )}{\partial k_{\omega, 1}}
\end{equation}
The chain rule is applied to the equation above
\begin{equation}\label{eq_lw_k1_gm2}
 \frac{\partial^{2} E_{\omega}}{\partial  k_{\omega,1}^{2} } = - \alpha_{\omega} \nu_{r} \omega_{r} \frac{\partial ( \ddot{e}_{2} + 2 \lambda_{\omega} \dot{e}_{2} + \lambda_{\omega}^{2} e_{2} )}{\partial \omega} \frac{\partial \omega}{\partial k_{\omega, 1}}
\end{equation}

First  $\frac{\partial  (\ddot{e}_{2} + 2 \lambda_{\omega} \dot{e}_{2} + \lambda^{2}_{\omega} e_{2})}{\partial \omega} =-\nu_{r}$  is obtained considering  \eqref{eq_errormodel2} and \eqref{eq_lateralerror2ndorder} and inserted into the equation above.  Then, total control input for the angular velocity applied to the mobile robot \eqref{eq_omega} and the feedforward control action for the angular velocity \eqref{eq_omegaf} are inserted into the equation above. Then, \eqref{eq_lw_k1_gm2} is obtained as follows:
\begin{eqnarray}\label{eq_lw_k1_gm3}
 \frac{\partial^{2} E_{\omega}}{\partial k_{\omega,1}^{2} } &=&  \alpha_{\omega} (\nu_{r})^{2} \omega_{r} \underbrace{ \frac{\partial ( \omega_{b} + \omega_{r} k_{\omega,1} + k_{\omega,0} )}{\partial k_{\omega,1} } }_{\omega_{r}} \nonumber \\
&=&  \alpha_{\omega} (\nu_{r})^{2} (\omega_{r})^{2}
\end{eqnarray}

The second derivative of the cost function $E_{\omega}$ with respect to the bias coefficient $k_{\omega,0}$ is computed by using the same procedure as follows:
\begin{eqnarray}\label{eq_lw_ko_gm}
\frac{\partial^{2} E_{\omega}}{\partial  k_{\omega,0}^{2} }  = \alpha_{\omega} (\nu_{r})^{2} 
\end{eqnarray}

Equations \eqref{eq_lw_k1_gm3} and \eqref{eq_lw_ko_gm} show that the sign of the curvature of the cost function for the angular velocity \eqref{eq_lw_ef} is always positive; therefore, there are no local minima, which indicate that the system reaches to the global minimum. After reaching the global minimum, since $\alpha_{\omega}$ is a constant, $v_{r}$ and  $\omega_{r}$ are bounded, the coefficient update algorithms \eqref{eq_lw_k1_5} and \eqref{eq_lw_k0} show that coefficients converge to a finite value. A finite value for the coefficients in steady-state results in a bounded feedforward control action \eqref{eq_omegaf}. 

\section{Experimental Validation}\label{sec_exp}

\subsection{Mobile Robot}\label{sec_mobilerobot}

\begin{figure}[t!]
  \centering
  \includegraphics[width=0.75\columnwidth]{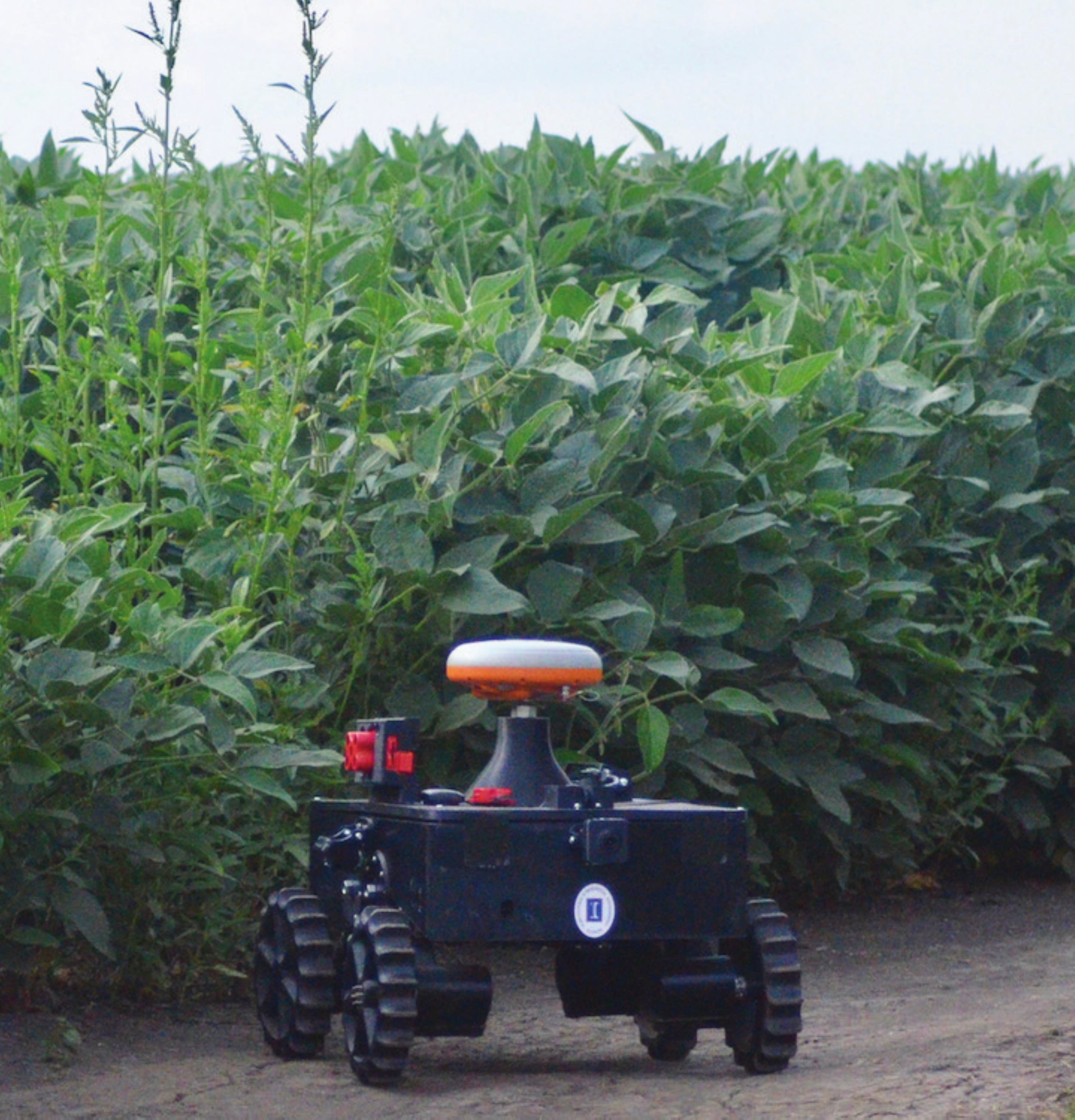} 
    \caption{The mobile robot, termed TerraSentia in off-road terrain.}\label{fig_mobilerobot}
\end{figure}

\begin{figure}[h!]
  \centering
  \includegraphics[width=1\columnwidth]{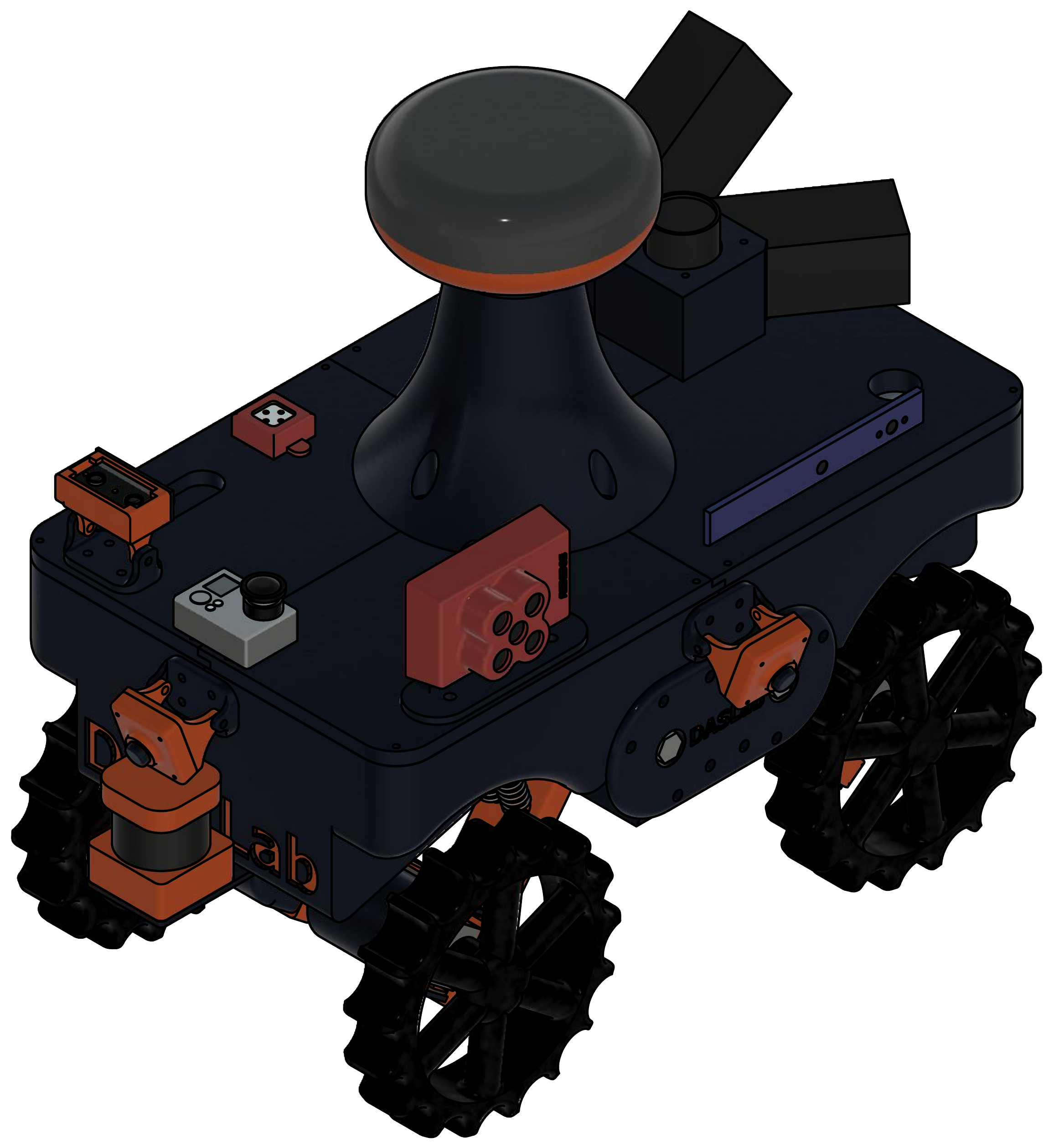} 
  \caption{CAD drawing of the ultra-compact 3D printed robot with a suite of sensors.}\label{fig_cad}
\end{figure}

The mobile robot, termed TerraSentia, is constructed entirely out of 3D printed construction as shown in Figs. \ref{fig_mobilerobot} and \ref{fig_cad}, which leads to an extremely light-weight robot (14.5 lbs) and has been proven to be structurally resilient to field conditions during an entire season of heavy operation in Corn, Sorghum, and Soybean farms in Illinois \cite{KayacanRSS}. We posit that this robot is an example of the potential of additive manufacturing (3D printing) in creating a new class of agricultural equipment that works in teams to replace, minimize, or augment traditional heavy farm equipment. In addition, lightweight equipment has several benefits, it is easier to manage, has better endurance, safer to operate in general, and leads to lower ownership cost.  On the other hand, the ultralight robot described here mitigates many of these challenges, yet, it leads to a uniquely challenging control problem in uneven and soil terrain in crop fields. The difficulties in control arise from complex and unknown wheel-terrain interaction and wheel slip. 

\begin{figure}[h!]
  \centering
  \includegraphics[width=1\columnwidth]{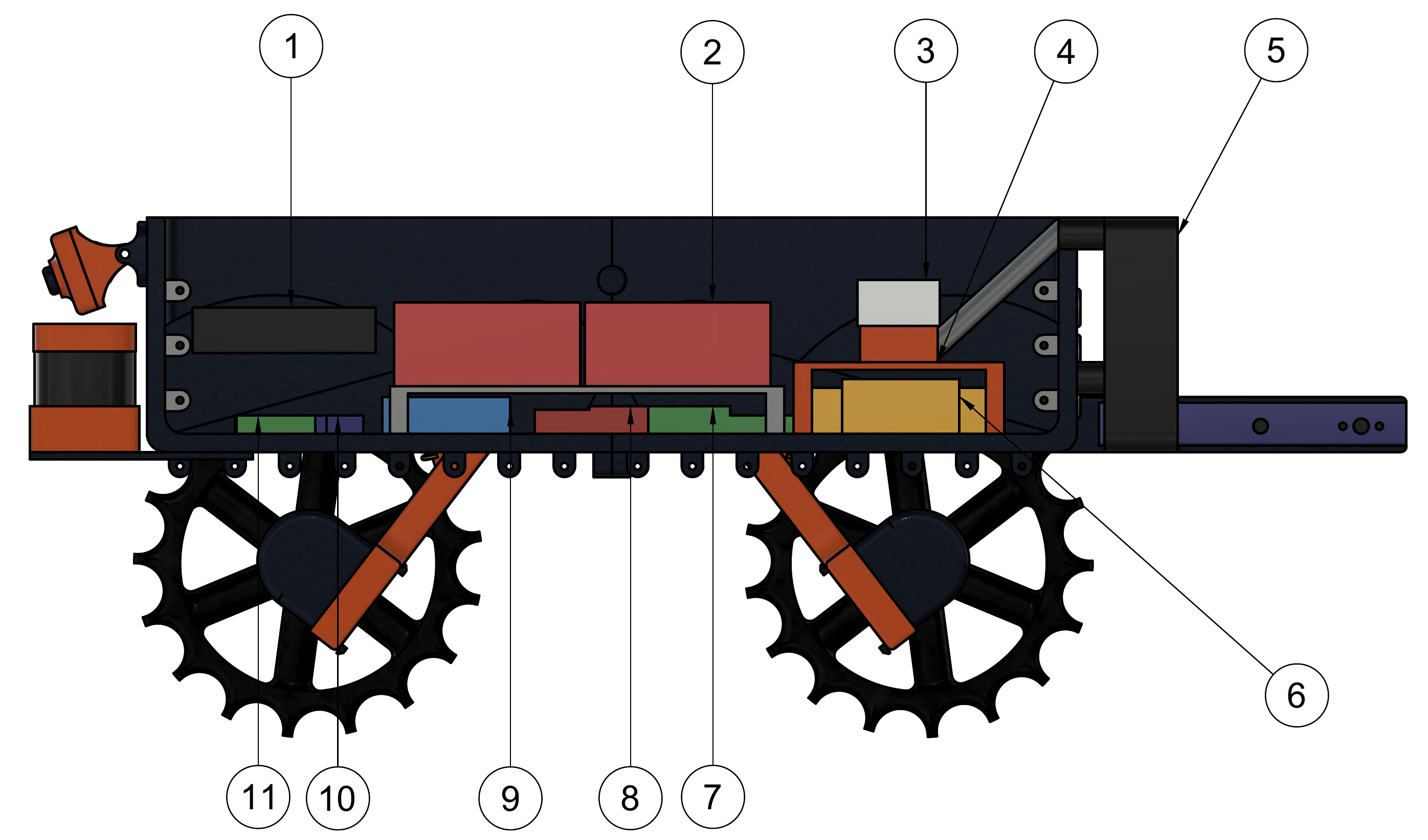} 
    \caption{Interior of the ultra-compact 3D printed robot. 1. Raspberry Pi, 2. Lithium Ion Batteries, 3. Tegra, 4. Heat sink, 5. Cooling Fan, 6. Kangaroo/Sabertooth, 7. Regulator, 8. 3-axis gyroscope, 9. Breadboard, 10. Raspberry Pi, 11. 3-axis accelerometer.}\label{fig_cad_interior}
\end{figure}

The placement of hardware is illustrated in Fig. \ref{fig_cad_interior}. One Global Navigation Satellite System (GNSS) antenna has been mounted straight up the center of TerraSentia, and the dual-frequency GPS-capable real-time kinematic differential GNSS module (Piksi Multi, Swift Navigation, USA) has been used to acquire centimeter-level accurate positional information at a rate of 5 Hz. Another antenna and module have been used as a portable base station and has transmitted differential corrections. There are four brushed 12V DC motors with a 131.25:1 metal gearbox (Pololu Corporation, USA), which are capable of driving an attached wheel at 80 revolutions per minute. A two-channel Hall-effect encoder (Pololu Corporation, USA) for each DC motor is attached to measure velocities of the wheels. The Sabertooth motor driver (Dimension engineering, USA) is a two-channel motor driver that uses digital control signals to drive two motors per channel (left and right channel) and has a nominal supply current of 12 A per channel. An onboard computer (1.2GHz, 64bit, quad-core Raspberry Pi 3 Model B CPU) acquires measurements from all available sensors and sends control signals to the Sabertooth motor driver in the form of two Pulse-width modulation signals. 

All available measurements from all its onboard sensors (GNSS and encoders) are fed to a state estimator, i.e., extended Kalman filter, to estimate heading angle of the mobile robot. In every time instant, estimates are fed to calculate the errors with respect to the inertial reference frame fixed to the motion ground and the estimated heading angle is fed to the transformation matrix to calculate the tracking errors in the frame on the mobile robot. Then, tracking errors are fed to the TELC algorithm, which generates the linear and angular velocity references to track the target path. Then, the linear and angular velocity references are controlled by a proportional-integral-derivative type motion controller - in other words, the robot's low-level controller - by using feedback from encoders attached to the motors to determine the required control signals in the form of the Pulse-width modulation signal. Thus, the tracking of given reference command signals ensures that the robot's desired velocities are maintained. The motion controller outputs the modified command signals to the Sabertooth Motor Driver which correlates the given control signals to the necessary output voltages needed by the DC motors.

\subsection{State Estimation}

An Extended Kalman Filter (EKF) is used for the state estimation in real-time because one GNSS antenna is not enough to obtain the heading angle of the mobile robot. We require heading angle information to calculate the heading angle error and also the rotation matrix. The inputs of the EKF are position information coming from the GNSS, and linear and angular velocities coming from the encoders and gyro. The outputs of the EKF are the position of the mobile robot on x- and y-coordinate system and the heading angle of the mobile robot. 

The discrete-time unicycle model used for the implementation of the EKF is written as follows:
\begin{eqnarray}\label{eq_kinematicmodelEKF}
x_{k+1} & = & x_{k} + T_{s} \nu_{k} \cos{\theta_{k}} \nonumber \\
y_{k+1} & = & y_{k} + T_{s} \nu_{k} \sin{\theta_{k}} \nonumber \\
\theta_{k+1} & = & \theta_{k} + T_{s} \omega_{k}
\end{eqnarray}
where $T_{s}$ is  the sampling interval. The general form of the estimated system model is written:
\begin{eqnarray}
\label{eq_generalmodelEKF}
\widehat{\textbf{q}}_{k+1} & = & f(\widehat{\textbf{q}}_{k}, \textbf{u}_{k}) + \textbf{w}_{k} \nonumber \\
\widehat{\textbf{z}}_{k+1} & = & h(\widehat{\textbf{q}}_{k}) + \textbf{v}_{k}
\end{eqnarray}
where $f(\widehat{\textbf{q}}_{k}, \textbf{u}_{k})$ is the system model \eqref{eq_kinematicmodelEKF}, $h(\widehat{\textbf{q}}_{k}) $ is the measurement function and $\textbf{z}_{k}=[x_{k}, y_{k}, \nu_{k}, \omega_{k}]^{T}$ is the measurements. The difference between the system model and real-time system is the process noise $\textbf{w}_{k}$ and observation noise $\textbf{v}_{k}$ in the measurement model. These noises are assumed to be independent and zero mean multivariate Gaussian noises with covariance matrices $\textbf{W}_{k}$ and $\textbf{V}_{k}$, respectively \cite{Goodarzi2017}:
\begin{eqnarray}\label{eq_noise}
\textbf{w}_{k} \backsim N(0,\textbf{W}_{k}) \nonumber \\
\textbf{v}_{k} \backsim N(0,\textbf{V}_{k})
\end{eqnarray}
where the weighting matrices are defined ad follows:
\begin{eqnarray}\label{eq_noise}
\textbf{W}_{k} &=& diag(0.1, 0.1, 0.1) \\
\textbf{V}_{k} &=& diag(0.03, 0.03, 0.01745)
\end{eqnarray}

\subsection{Experimental Results}

The experiment was carried out in off-road terrain and the GNSS is used as the ground truth measurements. The learning rate for the linear velocity $\alpha_{\nu} $ is set to $0.15$ and $0.05$ for the coefficients $k_{\nu,1}$ and $k_{\nu,0}$, respectively, while the learning rate for the angular velocity $\alpha_{\omega}$ is set to $0.1$ and $0.05$ for the coefficients $k_{\omega,1}$ and $k_{\omega,0}$, respectively. The positive constants $\lambda_{\nu}$ and $\lambda_{\omega}$ are set to $3$. The sampling time of the experiment is equal to $200$ milliseconds.

\begin{figure}[t!]
  \centering
  \includegraphics[width=1\columnwidth]{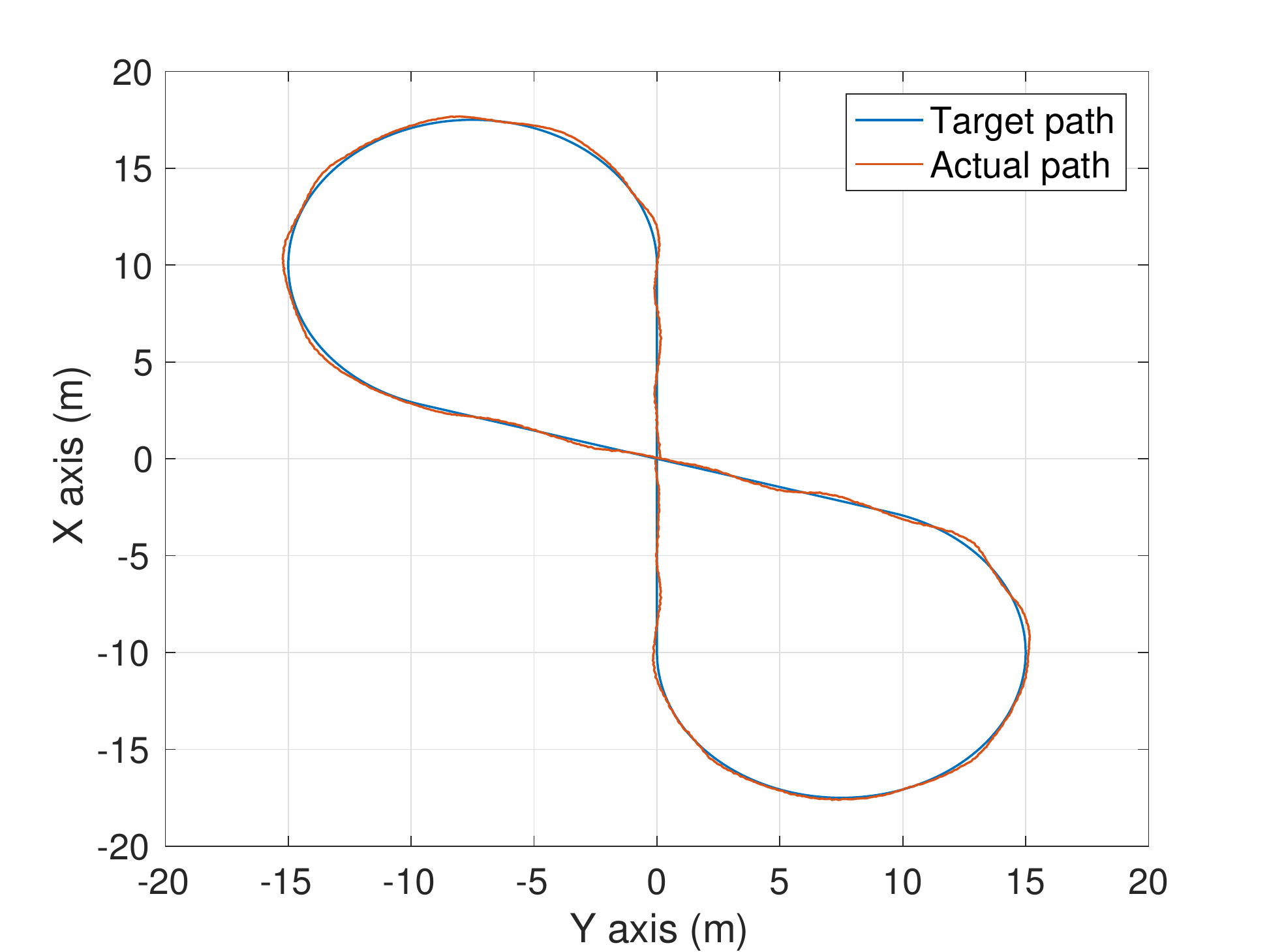}
  \caption{ Target and actual paths. Since an 8-shaped path consisting of two straight lines and two smooth curves was used to evaluate the path tracking performance of tracking-error model-based controllers in literature \cite{KLANCAR2007, BLAZIC20111, Kayacan2016, SKRJANC2017177}, we also use a similar 8-shaped path. The Euclidean error is plotted in Fig. \ref{fig_error}. }
  \label{fig_tra}
\end{figure}

An 8-shaped path is used as the reference path for the mobile robot to evaluate the path tracking performance of the TELC algorithm. The 8-shaped path consists of two straight lines and two smooth curves as illustrated in Fig. \ref{fig_tra}. The linear velocity reference is set to $0.3$ $m/s$ throughout the path generation. The angular velocity reference is zero for straight lines while it is set to  $\pm$ $0.05$ $rad/s$ for curved lines. Thus, the desired reference trajectory $x_{r}$, $y_{r}$ and $\theta_{r}$ is generated by considering the unicycle model in \eqref{eq_kinematicmodel} in the paper. The target and actual paths are shown in Fig. \ref{fig_tra}. The mobile robot controlled by the developed TELC algorithm can track the target path precisely.

\begin{figure}[t!]
  \centering
  \includegraphics[width=1\columnwidth]{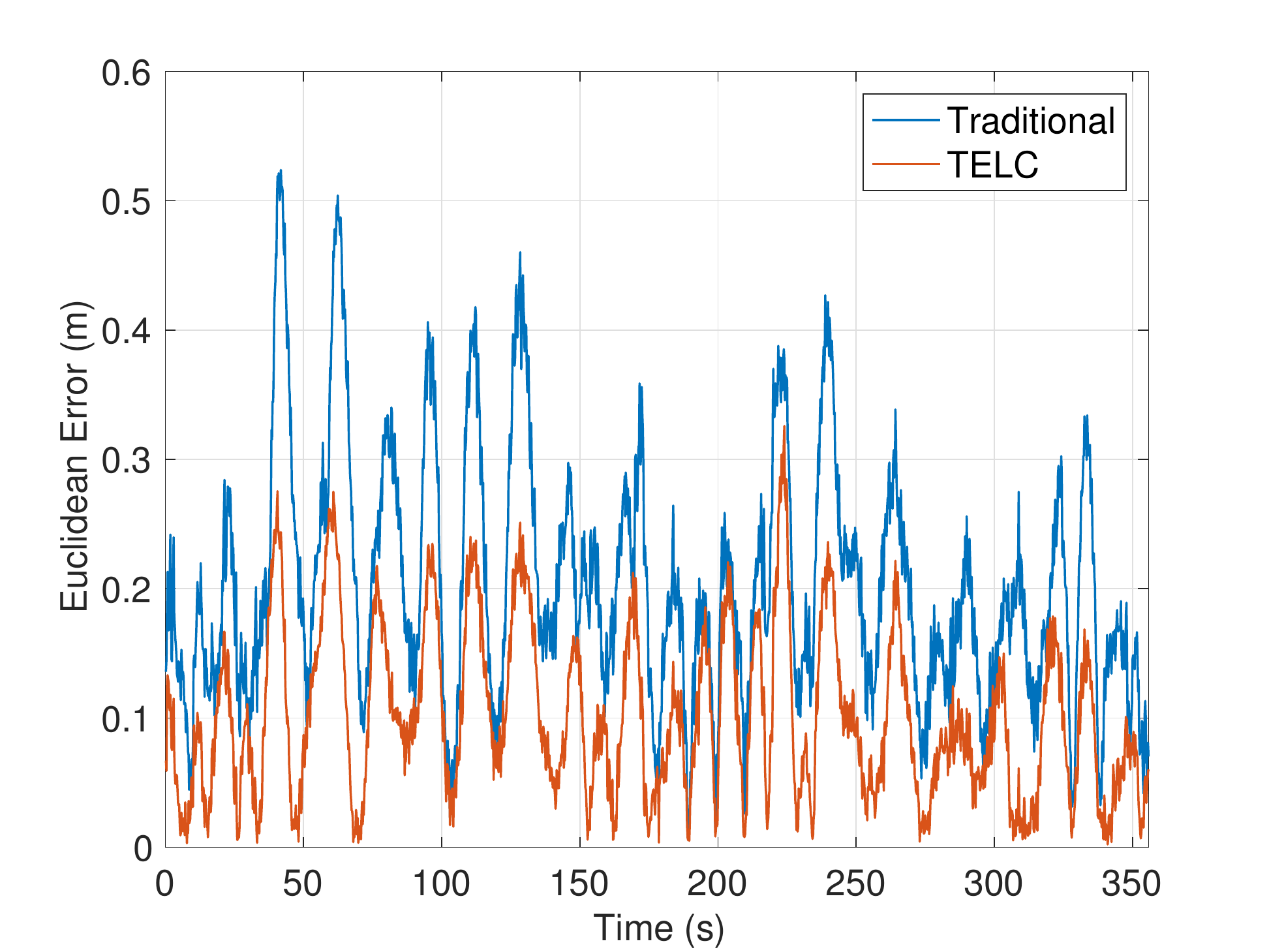}
  \caption{Euclidean error calculated using raw GNSS data. The mean values of Euclidean errors for the traditonal tracking error-based controller formulated in Section \ref{sec_tebc} and TELC algorithm are respectively around $20.31$ $cm$ and $9.11$ $cm$. This shows significant reduction on tracking error. }
  \label{fig_error}
\end{figure}

The Euclidean errors for the traditional tracking error control and TELC algorithms are shown in Fig. \ref{fig_error}. The mean values of the Euclidean errors for the traditional tracking error control formulated in Section \ref{sec_tebc} and TELC algorithms are respectively $20.31$ cm and $9.11$ cm. The TELC results in more precise mobile robot path tracking performance when compared to the traditional tracking error control method. This demonstrates the learning capability of the TELC algorithm in the outdoor environment.

\begin{figure}[h!]
  \centering
  \includegraphics[width=0.99\columnwidth]{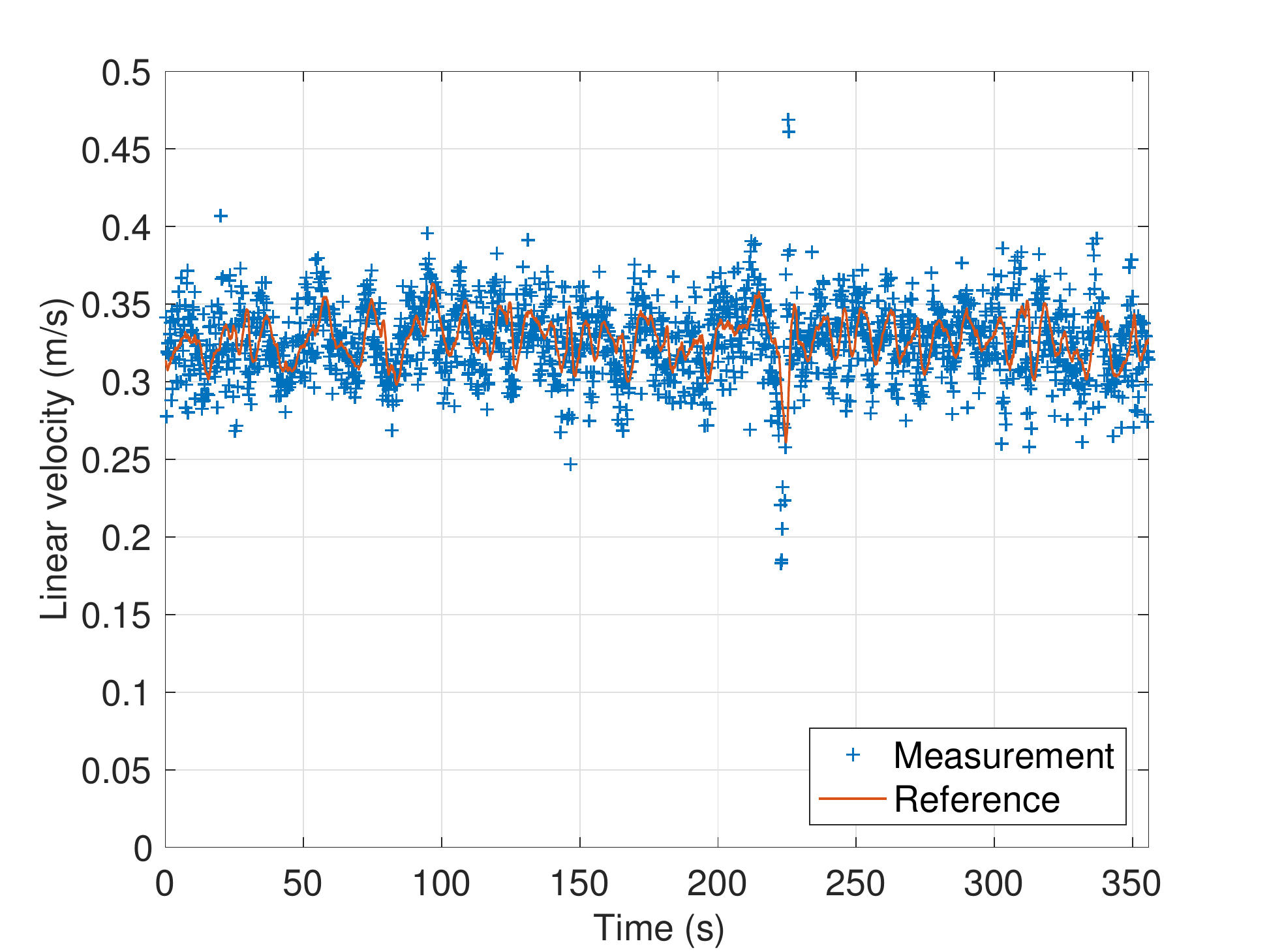}
  \caption{Reference and measured linear velocities. The low-level controller provides good tracking performance.}\label{fig_total_linear}
\end{figure}

\begin{figure}[h!]
  \centering
  \includegraphics[width=0.99\columnwidth]{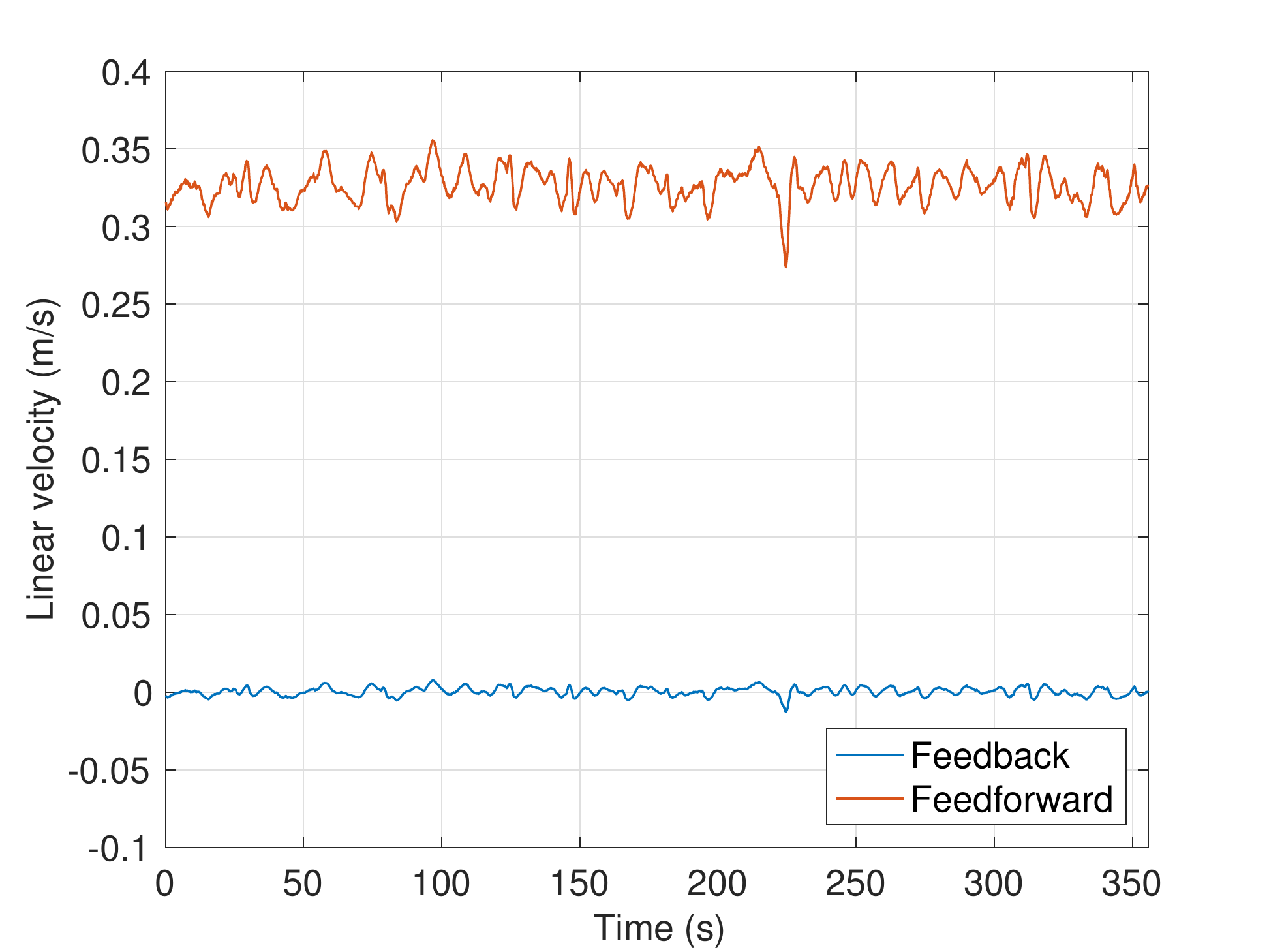}
  \caption{Control signals for the linear velocity. TELC algorithm learns the mobile robot dynamics so that the feedback control action is around zero while feedforward control action is updated and different than zero throughout the experiment.}\label{fig_fbff_linear}
\end{figure}

\begin{figure}[h!]
  \centering
  \includegraphics[width=0.99\columnwidth]{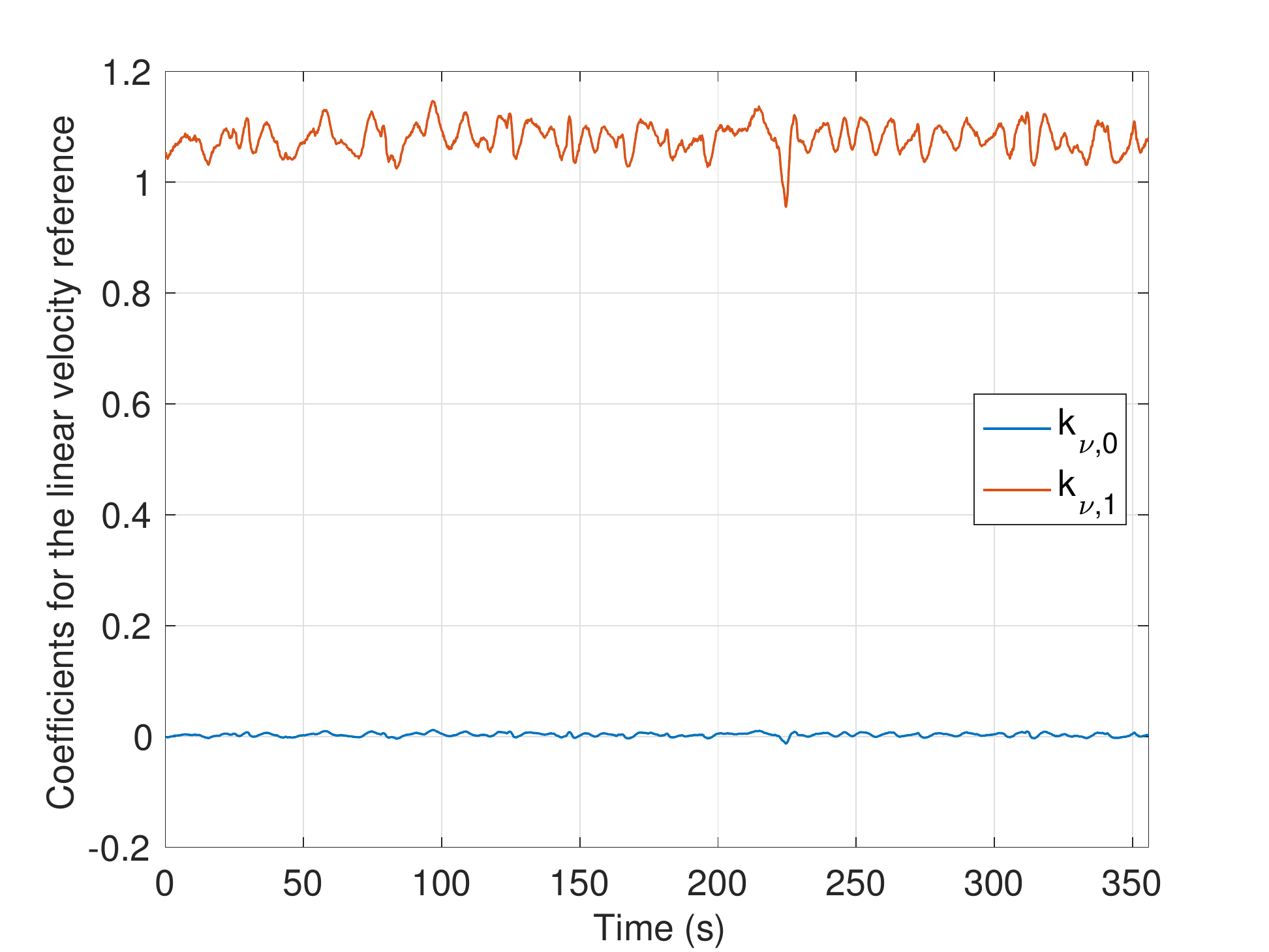}
  \caption{Adaptations of the coefficients for the linear velocity.}\label{fig_co_linear}
\end{figure}

The total linear velocity applied to the mobile robot is shown in Fig. \ref{fig_total_linear}. It is observed from the reference and measurements for the linear velocity that the low-level controller provides accurate tracking of the linear velocity reference despite the high noisy measurements. Moreover, the feedback control action generated by the MPC and the feedforward control action generated by the TELC algorithm are shown in Fig. \ref{fig_fbff_linear}. The feedback control action for the linear velocity is around zero as expected and the feedforward control action for the linear velocity takes the overall control action of the linear velocity of the mobile robot. Furthermore, the coefficients in the feedforward control action for the linear velocity are shown in Fig. \ref{fig_co_linear}. The nominal values for the coefficients $k_{\nu,1}$ and $k_{\nu,0}$ must be respectively $1$ and $0$ in the absence of uncertainties. These coefficients are different than the nominal values throughout experiments and varying due to the varying soil conditions in the outdoor environment. 


\begin{figure}[h!]
  \centering
  \includegraphics[width=0.99\columnwidth]{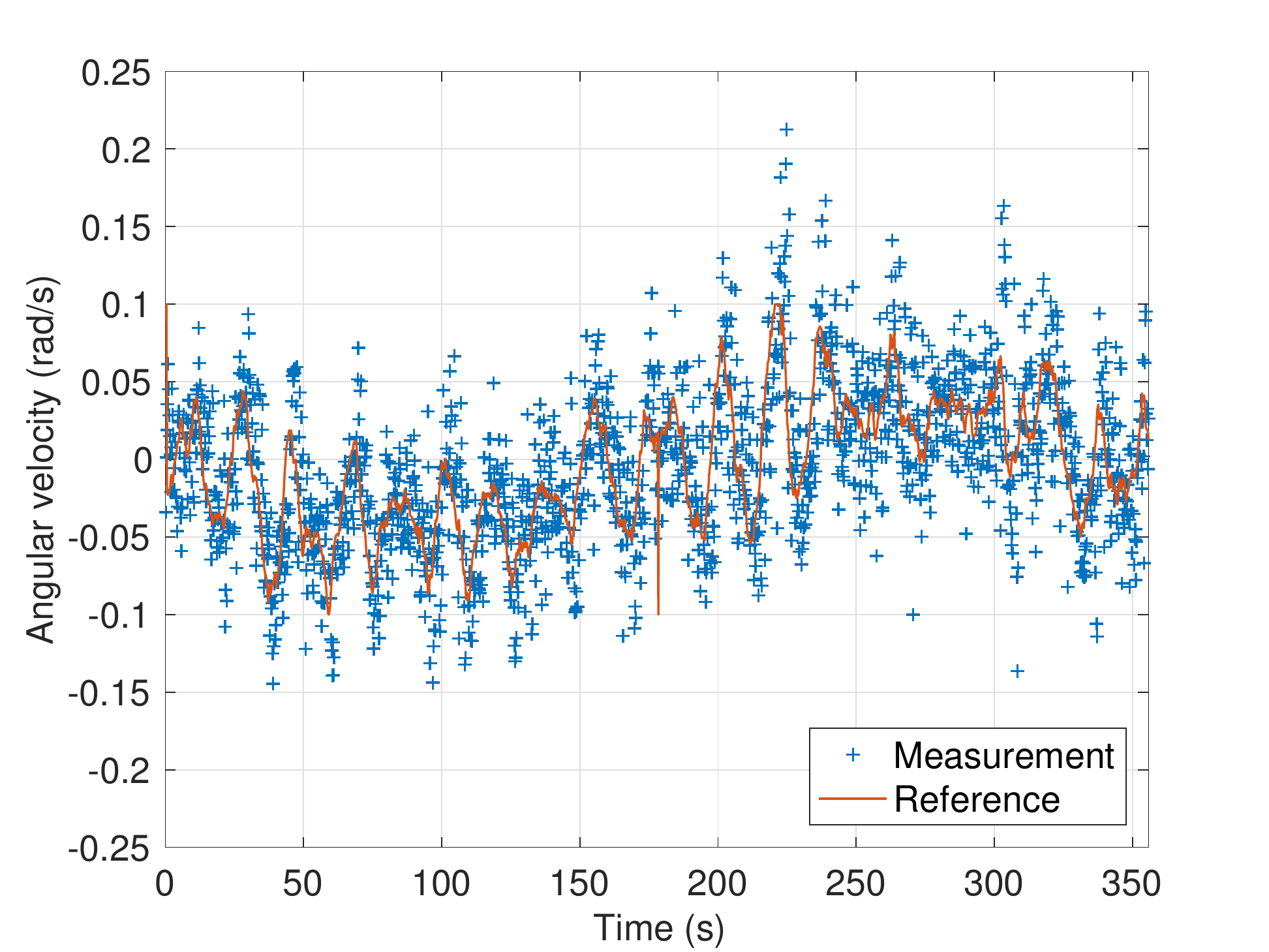}
  \caption{Reference and measured angular velocities. The low-level controller provides good tracking performance.}\label{fig_total_angular}
\end{figure}

\begin{figure}[h!]
  \centering
  \includegraphics[width=0.99\columnwidth]{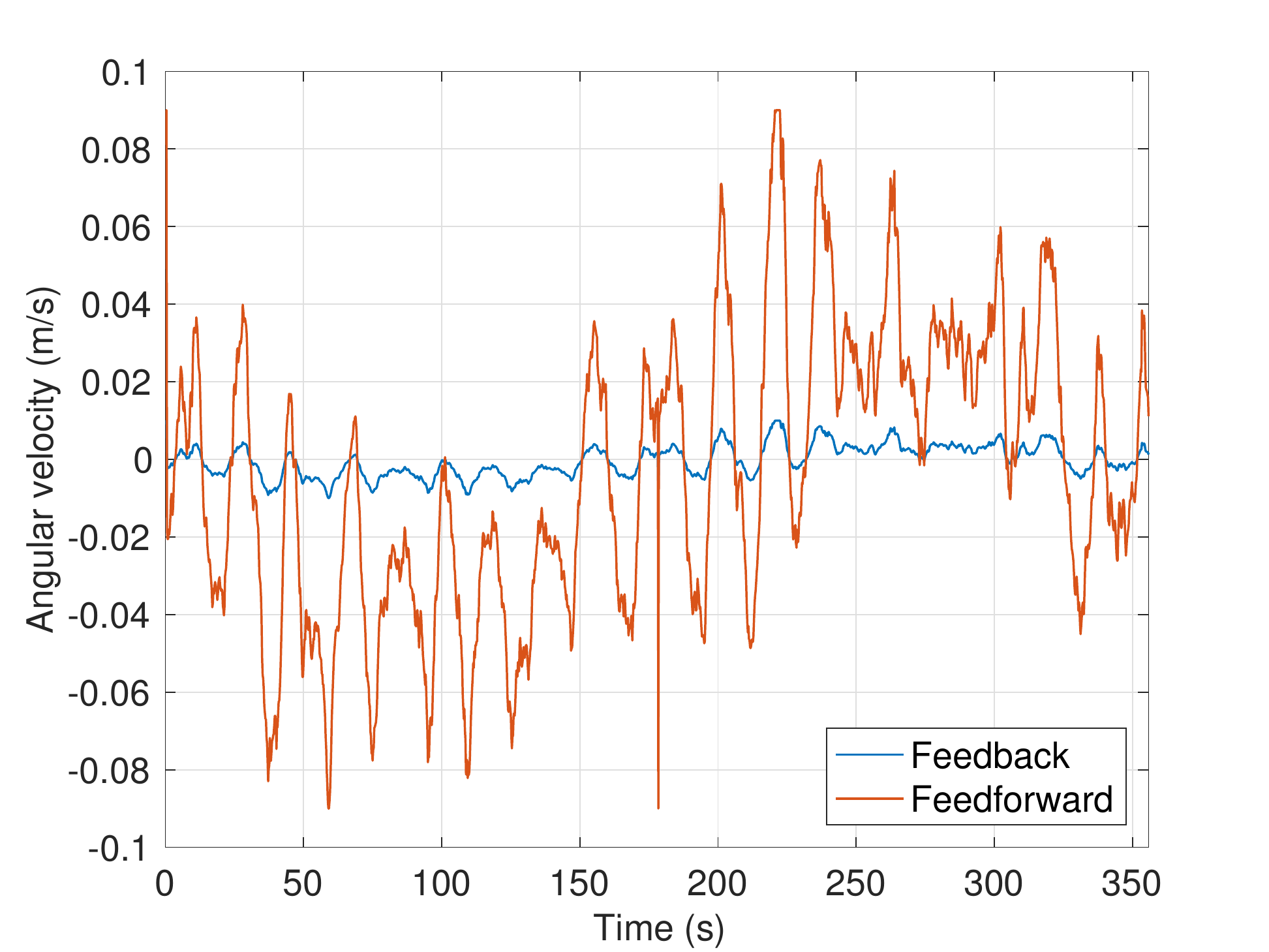}
  \caption{Control signals for the angular velocity. TELC algorithm learns the mobile robot dynamics so that the feedback control action is around zero while feedforward control action is updated and different than zero throughout the experiment.}\label{fig_fbff_angular}
\end{figure}

\begin{figure}[h!]
  \centering
  \includegraphics[width=0.99\columnwidth]{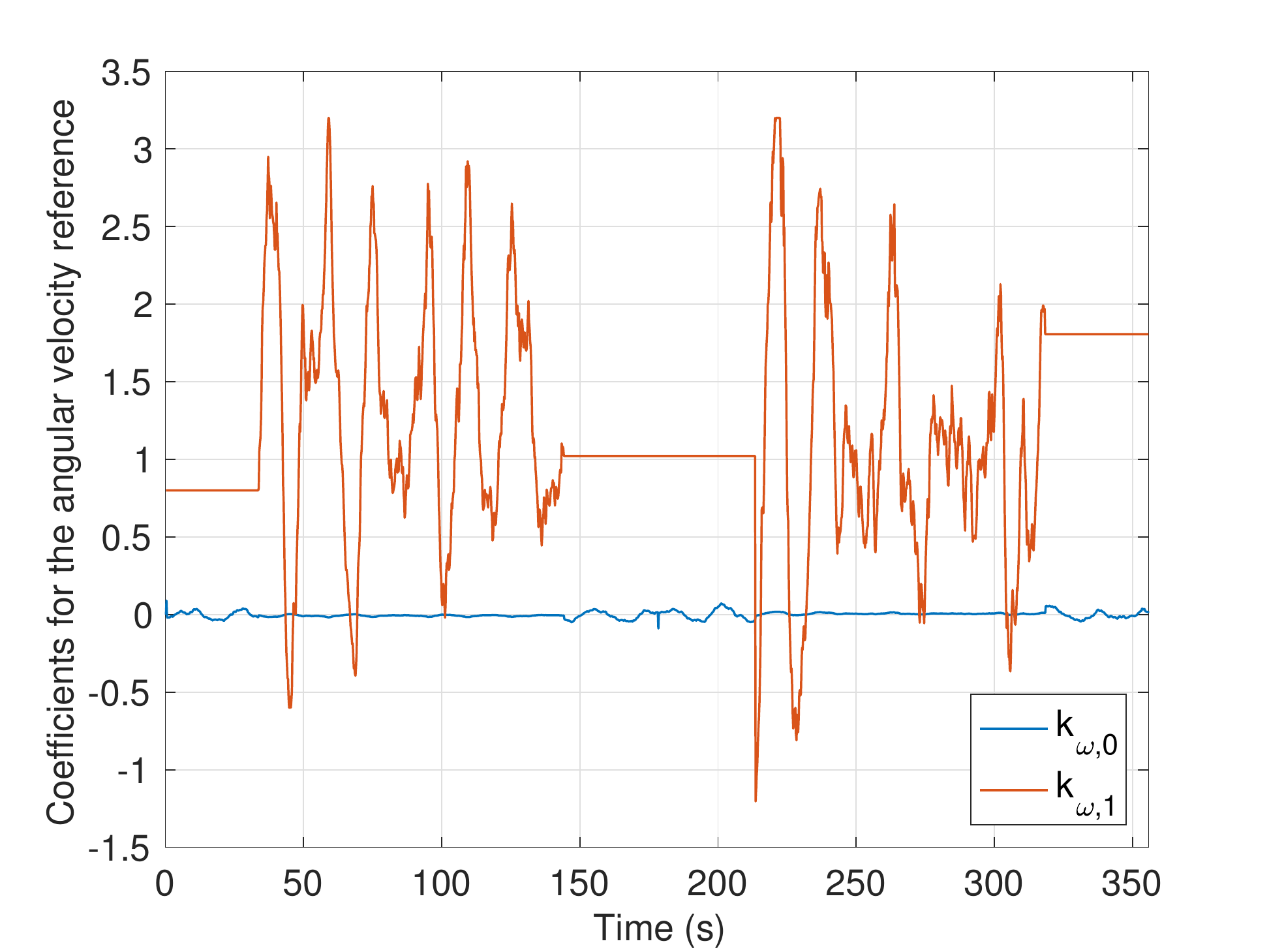}
  \caption{Adaptations of the coefficients for the angular velocity. }\label{fig_co_angular}
\end{figure}

The total angular velocity applied to the mobile robot is shown in Fig. \ref{fig_total_angular}. It is observed from the reference and measurements for the angular velocity that the low-level controller provides accurate tracking of the angular velocity reference despite the high noisy measurements. Moreover, the feedback control action generated by the MPC and the feedforward control action generated by the TELC algorithm are shown in Fig. \ref{fig_fbff_angular}. The feedback control action for the angular velocity is around zero as expected and the feedforward control action for the angular velocity takes the overall control action of the angular velocity of the mobile robot. Furthermore, the coefficients in the feedforward control action for the angular velocity are shown in Fig. \ref{fig_co_angular}. The nominal values for the coefficients $k_{\omega,1}$ and $k_{\omega,0}$ must be respectively $1$ and $0$ in the absence of uncertainties. These coefficients are different than the nominal values throughout experiments and varying due to the varying soil conditions in the outdoor environment. The angular velocity reference is equal to zero while the mobile robot is tracking straight lines. Therefore, the coefficient $k_{\omega,1}$ is constant, which can be seen from the update rule in \eqref{eq_lw_k1_5}. As a result, the only coefficient $k_{\omega,0}$ is updated to decrease the unmodeled effects in real-time.


\section{Conclusions}\label{sec_conc}

In this paper, a novel TELC algorithm has been developed for precise path tracking and experimentally validated on a mobile robot in the outdoor environment. In case of the plant-model mismatch, the TELC algorithm learns mobile robot dynamics by using the tracking error dynamics and updates the feedforward control actions. Therefore, the feedforward controller gradually eliminates the feedback controller from the control of the system once the mobile robot has been on-track. The experimental results on the mobile robot show that the TELC algorithm ensures precise path tracking performance as compared to the traditional tracking error-based control algorithm. The mean value of the Euclidean errors for TELC is 9 cm approximately in off-road terrain.

%

\bibliographystyle{spmpsci}      
\bibliography{ref_ttebmpc}   

\end{document}